\documentclass[10pt]{cai}

\graphicspath{{figs/}}

\usepackage{amsmath}

\begin{document}

\volumeheader{34}{0}
\begin{center}

  \title{Tracking agitation in people living with dementia in a care environment}
  
  \maketitle

  \thispagestyle{empty}

  \begin{tabular}{cc}
    Shehroz S. Khan\upstairs{\affilone\affilthree,*}, 
    Thaejaesh Sooriyakumaran\upstairs{\affiltwo},
    Katherine Rich\upstairs{\affilthree}
    Sofija Spasojevic\upstairs{\affilone}
    Bing Ye \upstairs{\affilthree}\\
    Kristine Newman \upstairs{\affilfour}
    Andrea Iaboni\upstairs{\affilone\affilthree}
    Alex Mihailidis\upstairs{\affilthree}
  \\[0.25ex]
  {\small \upstairs{\affilone} KITE, Toronto Rehabilitation Institute, University Health Network, Canada} \\
  {\small \upstairs{\affiltwo} McMaster University, Canada} \\
  {\small \upstairs{\affilthree} University of Toronto, Canada} \\
  {\small \upstairs{\affilfour} Ryerson University, Canada}
  \end{tabular}
  
  \emails{
    \upstairs{*}corresponding\_shehroz.khan@uhn.ca 
    }
  \vspace*{0.2in}
\end{center}

\begin{abstract}
Agitation is a symptom that communicates distress in people living with dementia (PwD), and that can place them and others at risk. In a long term care (LTC) environment, care staff track and document these symptoms as a way to detect when there has been a change in resident status to assess risk, and to monitor for response to interventions. However, this documentation can be time-consuming, and due to staffing constraints, episodes of agitation may go unobserved. This brings into question the reliability of these assessments, and presents an opportunity for technology to help track and monitor behavioural symptoms in dementia. 
In this paper, we present the outcomes of a $2$ year real-world study performed in a dementia unit, where a multi-modal wearable device was worn by $20$ PwD. In line with a commonly used clinical documentation tool, this large multi-modal time-series data was analyzed to track the presence of episodes of agitation in $8$-hour nursing shifts. 
The development of a baseline classification model (AUC=$0.717$) on this dataset and subsequent improvement (AUC= $0.779$) lays the groundwork for automating the process of annotating agitation events in nursing charts. 
\end{abstract}

\copyrightnotice

\section{Introduction}
\label{intro}
Many older adults in the advanced stages of dementia live in long-term care (LTC) or nursing homes settings to receive the needed care. In Canada, about one-third of PwD younger than $80$ years and $42\%$ above $80$ years live in long-term care homes  \cite{cihi}. 
However, this sector is poorly resourced and under-staffed, and this can impact on the ability of staff support the well-being of residents \cite{qualicare}.
A common challenge in these settings are the behavioral and psychological symptoms of dementia representing a heterogeneous group of non-cognitive type of symptoms and behaviours such as apathy, depression, irritability, agitation and anxiety \cite{cerejeira2012behavioral}. 
These behaviours are often in response to unmet needs, and their expression can place the residents and staff at risk.  Lack of staffing resources have an important impact on the ability to monitor the health status of residents.  With limited staff time and heavy workloads, the frequency, severity, and context around episodes of agitation are not always reliably documented \cite{cihi}.

Monitoring behavioural symptoms in dementia is clinically important. Changes in behaviour, such as increases in agitation, can signal a change in the health status of the resident, such as seen with delirium or worsening pain. Tracking changes in behaviour over time is also important when interventions are being trialed, to help determine their effectiveness. There is thus an important opportunity to use technology to develop objective measures of behavioural symptoms of dementia.

In this paper, we present the outcomes of a study performed at the Specialized Dementia Unit, Toronto Rehabilitation Institute, Canada. In this $2$-year study, $600$ days worth of wearable multi-modal sensor data was  collected from $20$ patients. Using this data, it has been shown previously that multi-modal wearable sensor data can be used to detect incidents of agitation in PwD with high accuracy \cite{khan2019agitation}. For this paper, we examine the predictive ability of this multi-modal sensor data to
replicate a common behaviour clinical documentation tool, the Pittsburgh Agitation Scale \cite{rosen1994pittsburgh}, by rating the presence or absence of agitation events in over an $8$ hour nursing shift. 
To approach this problem, we first reviewed nursing documentation and compared it to our research documentation of events of agitation. Next, we describe the data analysis and machine learning challenges associated with this problem, show baseline results and present improvements in achieving an Area Under the Curve (AUC) of Receiver Operating Characteristics (ROC) = $0.78$ on this unique dataset.

\section{Literature Review}
From a machine learning perspective, the problem studied in this paper can be termed as classification of long time-series data. This classification problem is a challenging task because of unreasonably high training / testing time and degenerating performance in the presence of noise. A common state-of-the-art benchmark is the 1-nearest-neighbor dynamic time warping (DTW) that many methods compare to \cite{susto2018time}. Sch{\"a}fer \cite{schafer2016scalable} presented a method based on bag-of-words approach. This method is suited for long-term time-series because it extracts subsequences and compares two time-series based on their structural similarity. The method is shown to be accurate, fast and robust to noise in comparison to DTW for both short and long time-series. Sharbiani et al. \cite{sharabiani2017efficient} presented an efficient dynamic time warping (DTW) technique by transforming the time-series into a set of segments with aggregated values and duration forming a reduced 3-D vector. The speed of the method is up to two orders of magnitude faster while providing similar performance for time-series of greater than $500$ points. Lahreche et al. \cite{lahreche2020fast} presented a method based on local-extrema and DTW for long time-series classification that is much faster and accurate than distance-based methods.

In the deep learning field, recurrent neural network (RNN) can be used for time-series classification. However, RNN suffers from the vanishing gradient problem due to training on long time-series. \cite{pascanu2013difficulty}. Time Warping Invariant Echo State Network has shown competitive accuracy on several long time-series datasets \cite{fawaz2019deep} as it can solve the vanishing gradient problem, especially when learning from long time-series.

Although the above stated research works are related, our problem is different due to the very large size of the time-series sequences (in the order of hundreds of thousands points) and the multi-modal nature of the sensor data (more details in Section \ref{sec:challenges}). Therefore, working directly on raw sensor points and applying approaches based on DTW are not feasible due to their quadratic time complexity. 
Our focus is to build a baseline model on this unique data, develop insights and investigate possible improvements. 

\section{Agitation Detection Study}
The data collection for this study took place between November 2017 - October 2019. Over this $2$-year study, $600$ days worth of wearable multi-modal sensor data was collected from $20$ participants who were people with dementia admitted to a Specialized Dementia Unit, Toronto Rehabilitation Institute, Toronto, Canada. The full protocol can be found at \cite{spasojevicpilot}; it is described in brief below.


\subsection{Data Collection Protocol}
An Empatica E4 wearable watch was placed on the dominant hand of the PwD. This watch can record data on the device that includes accelerometer, blood volume pulse, electrodermal activity and skin temperature. Fifteen cameras were also installed in the common areas of the unit to fine tune the annotations of specific agitation events. In this paper, we do not consider the video data, so it is not discussed further. Once a participant was identified by the geriatric psychiatrist, informed consent was obtained from their substitute decision-maker.
As per the study protocol, each participant was recruited in the study for a maximum of two months. If they did not show any symptoms of agitation for two consecutive weeks, they were removed from the study. The E4 device was removed from the participant’s wrist in the evening before bed and left to charge overnight. Therefore, the wearable sensor data for patients' sleeping time is not available. Every morning, the researcher would upload the data from E4 device to the cloud and then replace the E4 device on the participant’s wrist. 

The time, duration, and details of agitation events was determined through review of documentation in the nursing charts. Nurses were trained to record details of any agitation events and to flag these events by putting a green dot in the chart. These events were validated by review of research video footage from the unit.  In addition, nurses complete shift-by-shift clinical documentation of behavioural symptoms using the Pitsburgh Agitation Scale (PAS) \cite{rosen1994pittsburgh}.  For this study, we examined the PAS scores completed by the morning shift ($0700 - 1500$ hours) and evening shift ($1501 - 2300$ hours) as the E4 data for the overnight shift was not collected.  In practice, the E4 device was mostly placed on the patient after $0700$ hours in the morning and could be removed before $2300$ hours, if the participant retired to bed early. Therefore, for a given nursing shift, a full $8$ hour data may or may not be available for either of the morning or evening shifts.

\subsection{Clinical Documentation of Behavioural Symptoms}
\label{sec:pas}

The PAS \cite{rosen1994pittsburgh} is a clinical documentation tool used to record PwD’s severity of behavioural symptoms. The PAS is based on direct observation of PwD . The PAS rates the severity of agitation on the scale ranging from $0$ to $4$ in four behaviour group – Aberrant Vocalization (AV), Motor Agitation (MA), Aggression (AG) and Resisting Care (RC). Depending on needs, the period of observation could range from $1$ to $8$ hours. In this study, the PAS scores were completed at the end of each of the $8$-hour morning and evening nursing shifts. The overnight PAS scores were not used in this analysis because the corresponding multi-modal wearable sensor data is not available (as discussed above).

When examining the PAS scores completed for the study participants, we noticed some common issues.  There were missing data for PAS scores ($14.72\%$ for AV and AG, and $15.58\%$ for MA and RC).  We also discovered that many shifts in which agitation had been recorded in the research data were scored as absent of agitation.  
To understand the distribution of PAS scores reported in the nursing charts, we plotted the scores for each type (AV, MA, AG, RC) for both the known agitation and non-agitation shifts based on our research data collection. An agitation shift is defined as the one that contained at least one mention of agitation event in the nursing chart. Similarly, a non-agitation shift is defined as the one that contained no mention of agitation event in the nursing chart . The general expectation is that for agitation shifts there will be more PAS scores with higher values and for non-agitation shifts there will be more PAS scores with lower values. 

\begin{figure}[ht!]
    \centering
    \includegraphics[scale=0.47]{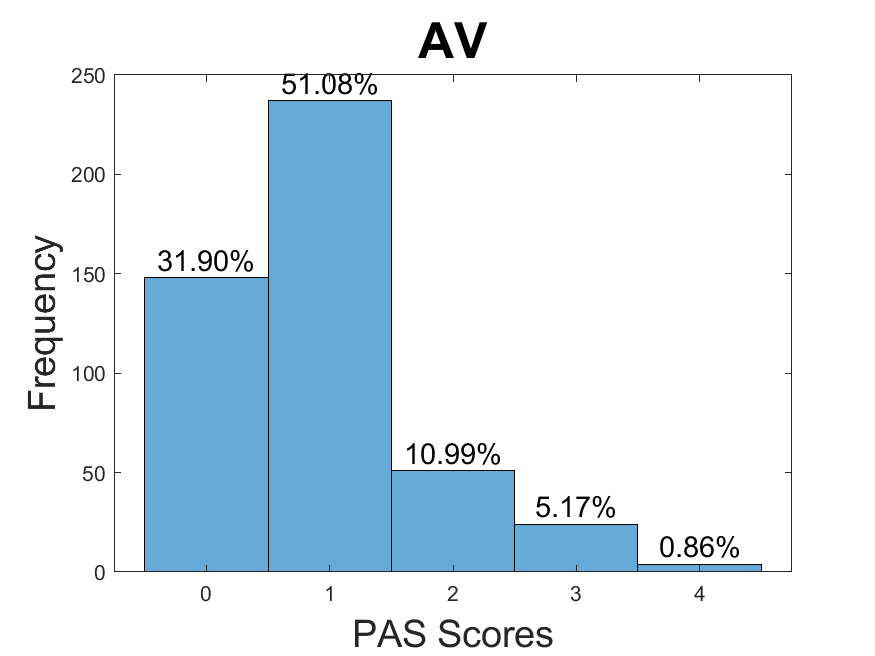}\hfill
    \includegraphics[scale=0.47]{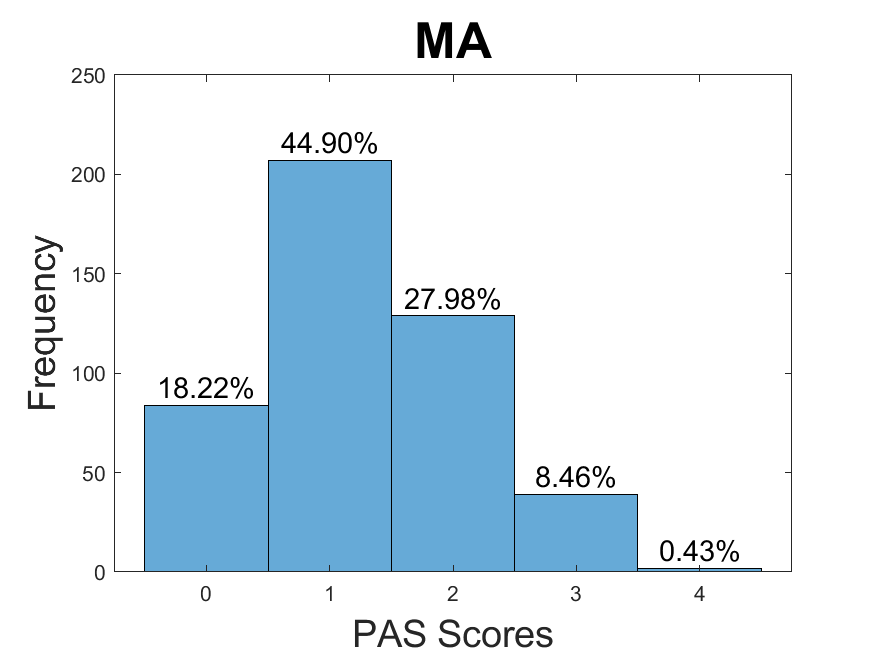}\\
    \includegraphics[scale=0.47]{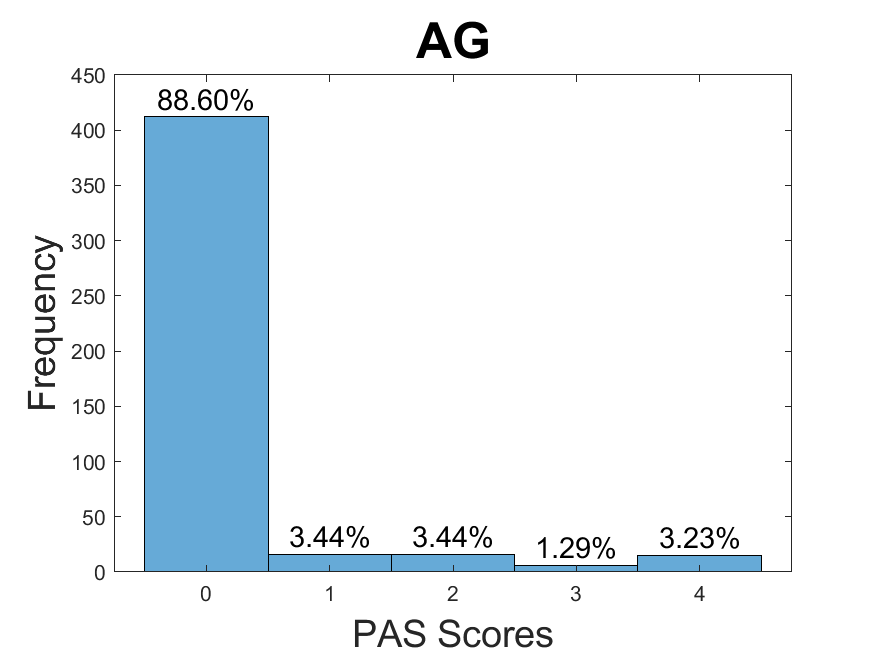}\hfill
    \includegraphics[scale=0.47]{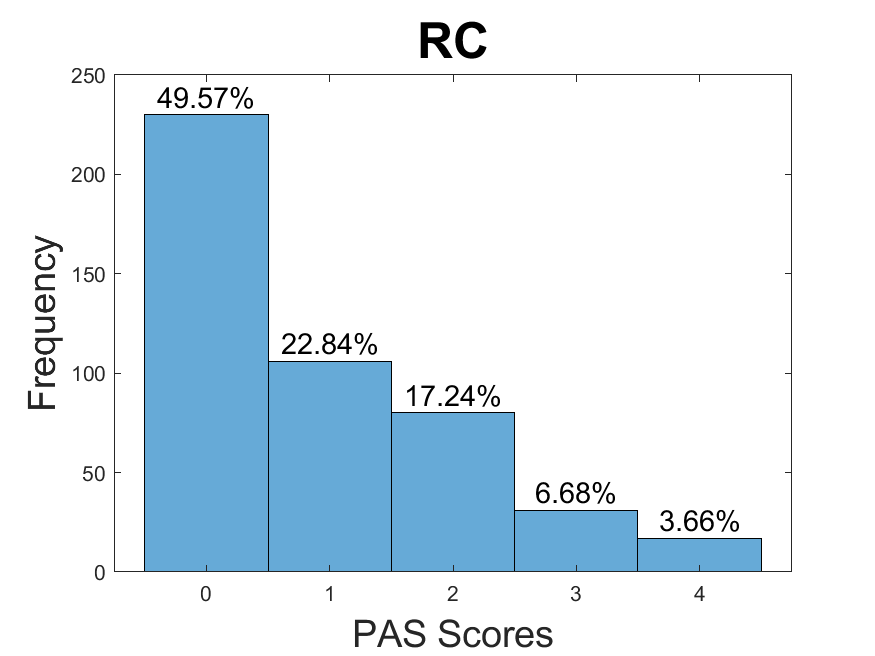}
    \caption{Distribution of PAS scores on non-agitation shifts.}
    \label{fig:pasnonag}
\end{figure}

\begin{figure}[ht!]
    \centering
    \includegraphics[scale=0.47]{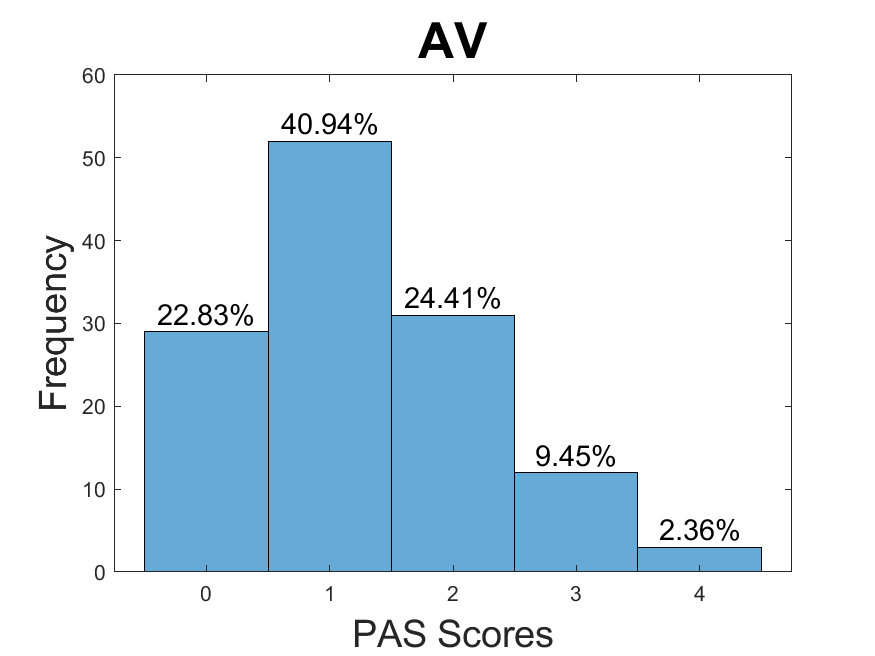}\hfill
    \includegraphics[scale=0.47]{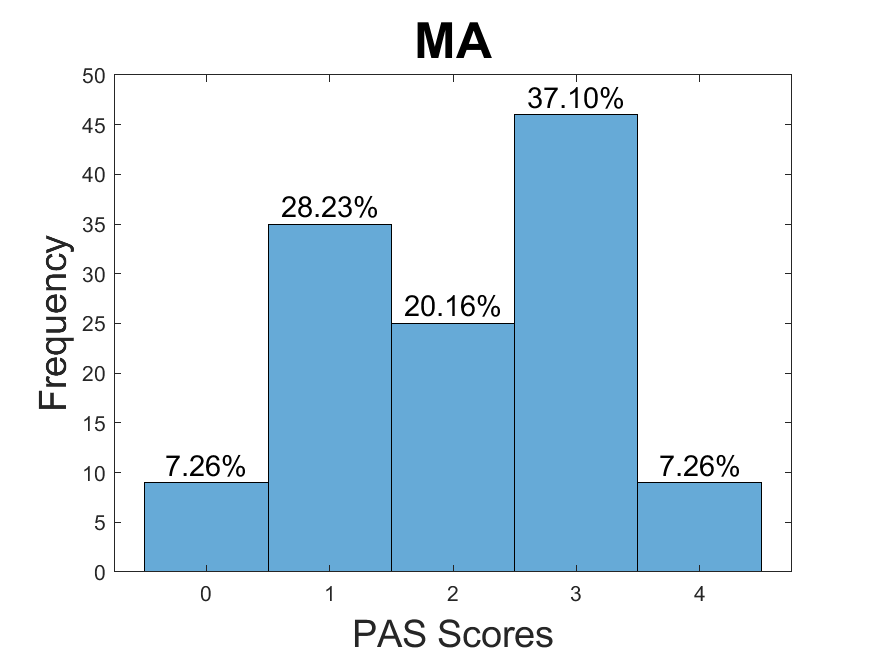}\\
    \includegraphics[scale=0.47]{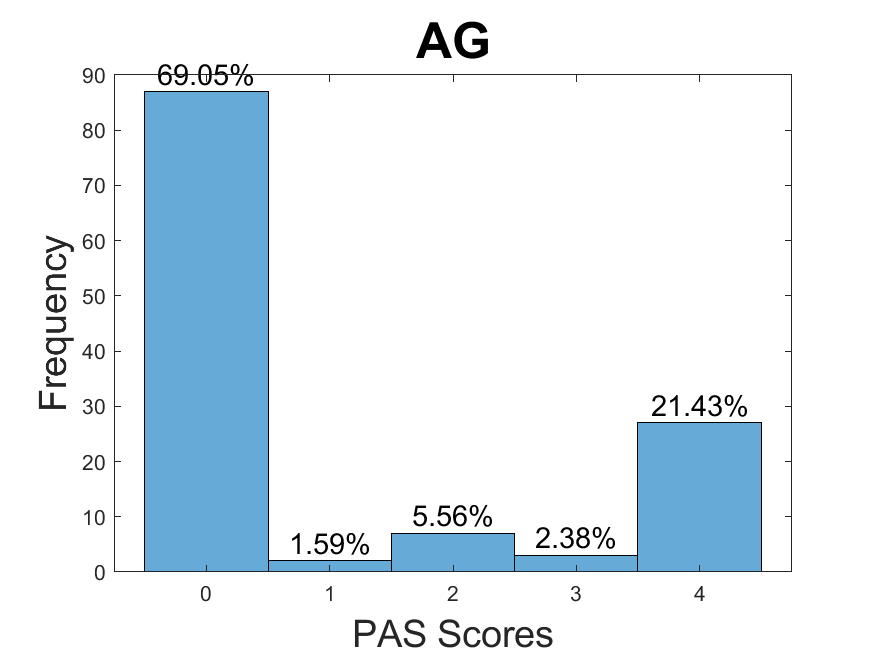}\hfill
    \includegraphics[scale=0.47]{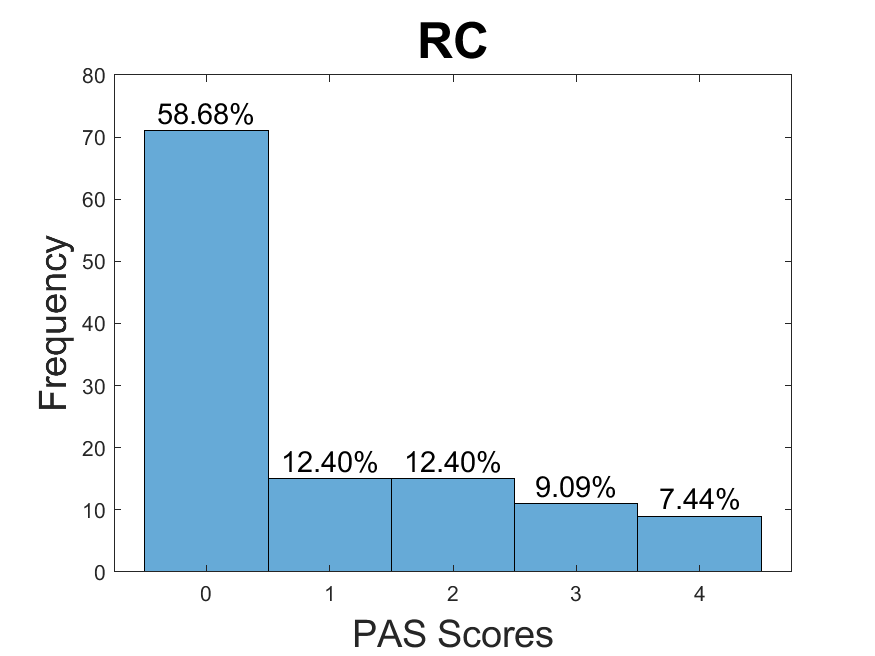}
    \caption{Distribution of PAS scores on agitation shifts.}
    \label{fig:pasag}
\end{figure}

Figure \ref{fig:pasnonag} shows the histogram of PAS scores for AV, MA, AG and RC for non-agitation shifts. As expected, majority of the scores are below or equal to $3$. This indicates that the patients were mostly calm in the non-agitation shifts. Figure \ref{fig:pasag} shows the histogram of PAS scores for AV, MA, AG and RC for agitation shifts. We observed that for AV, AG and RC the PAS scores were on the lower side, whereas for MA ($\approx 56\%$) of the score was less than $3$. This indicated a surprising contrast that on a shift when agitation occurred (and noted in the nursing chart), the PAS scores given may not be high (e.g. more than $3$). This under-reporting of agitation events in the clinical documentation gave rise to our research question. 

The focus of this analysis is thus to improve the accuracy of documentation of behavioural symptoms on a shift-by-shift basis, by using sensor data to classify each shift as agiation or no agitation.  To answer this question, we used as our gold standard annotation, the presence (or absence) of agitation on any given shift which was collected from a detailed review of nursing charts and validated using research video review. 

\subsection{Machine Learning Challenges}
\label{sec:challenges}
The data collected in this study poses several machine learning challenges:
\begin{itemize}
    \item The agitation label is only available after a long shift or sequence of multi-modal sensor data. In an ideal case, the shift should be $8$ hours. The sensors are sampled at $64$Hz (more details in Section \ref{sec:dataprocessing}); therefore, an $8$ hour shift will result in $64 \times 60 \times 60 \times 8 = 1,843,200$ points per sensor. There are four sensors in the Empatica E4 watch -- accelerometer, blood volume pulse, electrodermal activity and skin temperature. These points are followed by one label -- $0$ for absence and $1$ for presence of agitation in that shift. We observed sequences less than $8$ hours as well (see Section \ref{sec:improve}); however, the order of data remains similar.
    
    \item The duration of agitation events can have important effect in terms of extracting relevant features from the long multi-modal sensor data sequence. Previous work suggests that agitation events can vary from one minute to up to three hours \cite{spasojevicpilot}.
    However, the actual start/end time or duration of an agitation event is not available for this problem. 
    
    \item The length of the multi-modal sensor sequences is not the same. The primary reason is that the sensors may not capture the entire data corresponding to the $8$-hour nursing shift (see more details in Section \ref{sec:improve}). Therefore, building any temporal predictive model or strategy for feature extraction is not straight forward.
\end{itemize}

\section{Data Processing and Experimental Setup}
\label{sec:dataprocessing}
Different sensors in the Empatica E4 device sample data at a different sampling frequency (Accelerometer at 32Hz, Blood Volume Pulse at 64Hz, Electrodermal Activity and Skin Temperature at 4Hz). To avoid data loss, all the sensors were re-sampled to 64Hz to match with the maximum sampling rate of Blood Volume Pulse \cite{spasojevicpilot}. Then, the multi-modal sensor data is split into shifts based on the timestamps (available with the sensor data)  -- morning shift ($0700 - 1500$ hours) and evening shift ($1501 - 2300$ hours). In total, $693$ shifts were extracted that contained sensor data. Then, for each shift the following $10$ features \cite{khan2014x} were extracted from each of the sensor modality -- mean, minimum, maximum, standard deviation, interquartile range, difference of maximum and minimum, number of abrupt changes in the data, greatest abrupt change in the data, greatest gradient value in the data and coefficient of variation (mean divided by standard deviation). Additionally, for electrodermal activity, phasic and tonic signal were extracted.  Four features each were extracted for tonic and phasic signals --  trapezoidal numerical integration, total number of peaks, their maximum and minimum. Two more features were extracted from tonic signal -- maximum of difference and approximate derivative, and mode of the signal. In some cases, the coefficient of variation for temperature gave a $NaN$ values. This could happen when the standard deviation is zero, meaning no change in skin temperature during the recording session. Therefore, to avoid numerical computation problems, this feature was removed and overall $49$ features were considered for each shift for classification purposes.

We trained three classifiers to detect the presence or absence of agitation in a given shift -- Logistic Regression (LR), Random Forest (RF) and Support Vector Machine (SVM). We performed $10$-fold cross validation. An internal $5$-fold cross validation is performed to tune the parameters of these classifiers. The tuned parameters for LR was lambda (regularization term) in the range of $[0.01, 0.05, 0.1, 1, 10, 100]$, for RF were Number of Trees ($[10,30,50,70,90]$) and Number of Predictors ($[f/5, 2*f/5, , 3*f/5, 4*f/5]$, where $f$ is the number of features), and for SVM were Box Constraint and Kernel Scale (both in the range $[0.01, 0.1, 1, 10, 100,1000]$).

Out of the $693$ sensors shifts, $140$ contained agitation, i.e. approximately $20\%$ contained agitation shifts and rest of them did not. Clearly, the data labels are skewed that may favour the majority class. Therefore, we introduced a misclassification `cost'  parameter in the RF, LR and SVM classifiers. The value of cost is calculated within each fold of the cross-validation step as follows. Let $L$ be a variable that contains the $N$ labels $\in (0,1)$, where $0$ means non-agitation shift and $1$ means agitation shift, and $w$ is the ratio of total number of agitation shifts and total number of shifts, defined as:
\begin{equation}
    w = \frac{\sum_i^N (L_i \neq 0)}{\sum_i^N L_i}
\end{equation}

and the cost matrix is defined as:
\begin{equation}
    cost =    
\begin{bmatrix}
    0 &  \frac{1}{(1-w)} \\
    \frac{1}{w} & 0 \\ 
\end{bmatrix}
\end{equation}
This cost matrix shows that when a minority class, i.e. agitation is predicted wrongly as non-agitation it will be penalized heavily as $\frac{1}{w}$ in comparison to when a non-agitation is wrongly classified as agitation $\frac{1}{1-w}$ (considering $w<0.5$)
It should be noted that in a classifier without additional misclassification cost, the off-diagonal elements are $1$. The folds in each cross-validation step were stratified, so that data from both majority (non-agitation) and minority (agitation) class is spread equally in each fold. Since the cross-validation is stratified, the cost parameter will remain almost constant in each fold. During each fold of the cross-validation, the  scores/probabilities on the test set are concatenated. The final concatenated vector of scores is used to calculate the area under the curve (AUC) of the receiver operating characteristic (ROC), which is the reported performance metric of different classifiers. The AUC of a random classifier is $0.5$ and for the best classifier is $1$. A higher value of AUC means more confidence is detecting the class of interest (agitation in our case).

\begin{table}[ht!]
    \centering
    \begin{tabular}{|c|c|} \hline
        \textbf{Classifier} & \textbf{AUC} \\ \hline
        RF &  0.688 \\ \hline
        RF$_{cost}$ & 0.690 \\ \hline
        LR & 0.717 \\ \hline
        LR$_{cost}$ & 0.705 \\ \hline
        SVM & 0.601 \\ \hline
        SVM$_{cost}$ & 0.699 \\ \hline
    \end{tabular}
    \caption{Performance of different classifiers on baseline data, both with and without cost.}
    \label{tab:baseline}
\end{table}

\section{Results}

\subsection{Baseline Model}
\label{sec:baseline}
Table \ref{tab:baseline} shows the classification results on each of the classifier without and with cost (with a subscript). We observe that including cost in the classification slightly improves the AUC in both RF and SVM classifiers (shown as RF$_{cost}$ and SVM$_{cost}$) ; however, not so much in the case of LR. Although, LR without cost gives highest AUC, the AUCs of RF$_{cost}$ and SVM$_{cost}$ classifiers are also very similar.

\subsection{Improving the Baseline}
\label{sec:improve}

As mentioned earlier, a typical nursing shift is $8$ hours long. However, there can be situations where the sensor data collected for those shifts may be less than $8$ hours. Some of these scenarios are: 
\begin{itemize}
    \item The watch was attached to the patient late in the morning shift (i.e. not at $0700$ hours).
    \item The watch was removed from the patient before the shift ends (i.e. before $2300$ hours) or anytime before that).
    \item The data collection was stopped either by the patient or nursing staff during the daily care (e.g. bathing).
\end{itemize}

Therefore, we define \textit{sensors shift} as the time duration for which the sensor data is available to the corresponding nursing shift. If the sensor shift contains $8$ hours of uninterrupted data, then it is equivalent to a nursing shift.

\begin{figure}[ht!]
    \centering
    \includegraphics[scale=0.45]{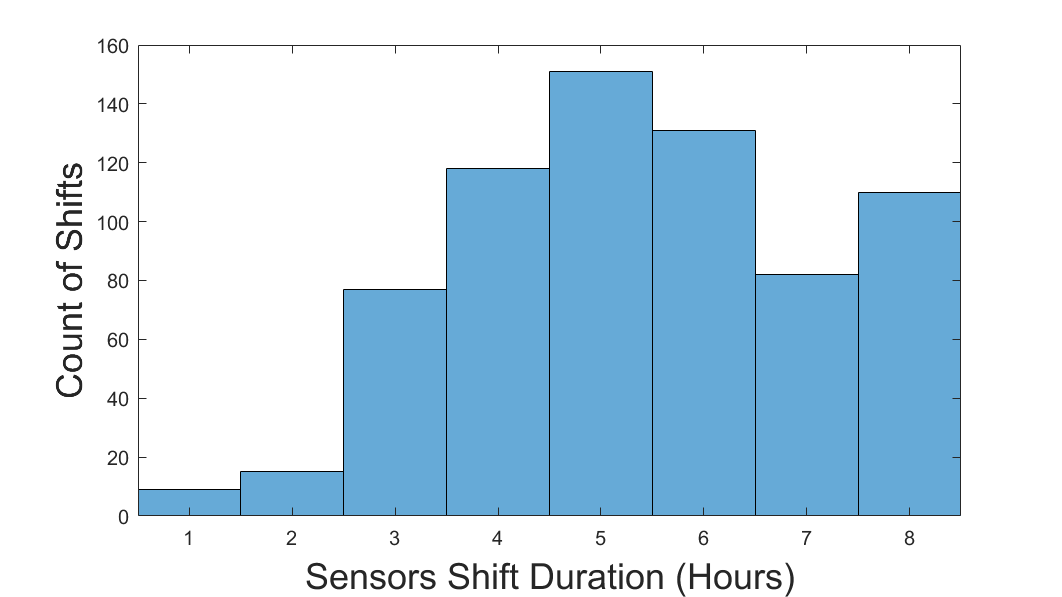}
    \caption{Distribution of multi-modal sensor data across sensors shifts of various duration.}
    \label{fig:shiftbins}
\end{figure}

We plot a histogram to verify the distribution of multi-modal sensor data across various shifts. Figure \ref{fig:shiftbins} shows the distribution of multi-modal sensor data corresponding to sensors shifts of different sizes. The histogram clearly shows that majority of the sensor data does not correspond to the usual $8$ hours nursing shift because of the reasons discussed above. 
Table \ref{tab:shiftbins} shows the distribution of sensors shift data across different bins, where Bin$x$ means it contains $x$ hours of multi-modal sensor data.
The first row of Table \ref{tab:shiftbins} shows that out of total $693$ shifts, only $110$ shifts have data corresponding to $8$ hours nursing shift (approximately $15.87\%$) and rest of the sensor data is distributed in sensors shifts of duration $1$ to $7$ hours. This imposes another challenge that the multi-modal sensor data is not equal in length; hence feature extraction becomes non-trivial. Another observation was that the distribution of agitation shifts in these sensors shifts is almost similar (third row of Table \ref{tab:shiftbins}).

\begin{table}[ht!]
    \centering
    \begin{tabular}{|c|c|c|c|c|c|c|c|c|} \hline
          & Bin1 & Bin2  & Bin3  & Bin4  & Bin5  & Bin6  & Bin7  & Bin8 \\ \hline
         All Shifts &  9  &  15 & 77 &  118 &  151 &  131 &  82 &  110 \\ \hline
         Agitation Shifts & 2 & 3 & 17 & 25 & 37 & 23 & 13 & 20 \\ \hline
         \% of Agitation Shifts & 22.22 & 20 & 22.08 & 21.19 & 24.50 & 17.56 & 15.85 & 18.18 \\ \hline
    \end{tabular}
    \caption{Distribution of multi-modal sensor data with respect to different shifts.}
    \label{tab:shiftbins}
\end{table}

In this dataset, a very long multi-modal sensor sequence is given, the only information available is whether agitation event occurred or not. The actual timing and duration of the agitation events are not provided. Therefore, the features extracted must reflect the occurrence of agitation. In a very large sequence (e.g. $8$ hours sensors shift), it is very challenging to extract meaningful features without knowing their actual timing and number of occurrences during a nursing shift. The larger the sequence is, the more difficult is to extract discriminatory features. Therefore, we hypothesize that considering shorter sequences for feature extraction will be more useful than larger sequences. To test this hypothesis, we performed the following two experiments:
\begin{enumerate}
    \item Larger Sequences Experiment (LSE) -- In this experiment, we only consider the sequences that are larger than a specific threshold, s.t. $SequenceLength \ge b$, where $b$ is varied from $0$ to $4$ hours (or $240$ minutes) with an equal increments of $15$ minutes. Therefore, LSE will result in selecting sequences of larger length from the overall sensor data.  
    \item Smaller Sequences Experiment (SSE) -- In this experiment, we only consider the sequences that are smaller than a specific threshold, s.t. $SequenceLength \le b$, where $b$ is varied from $8$ to $4$ hours with an equal decrements of $15$ minutes. Therefore, SSE will result in selecting sequences of smaller length from the overall sensor data.
\end{enumerate}

\begin{figure}[!ht]
    \centering
    \includegraphics[scale=0.4]{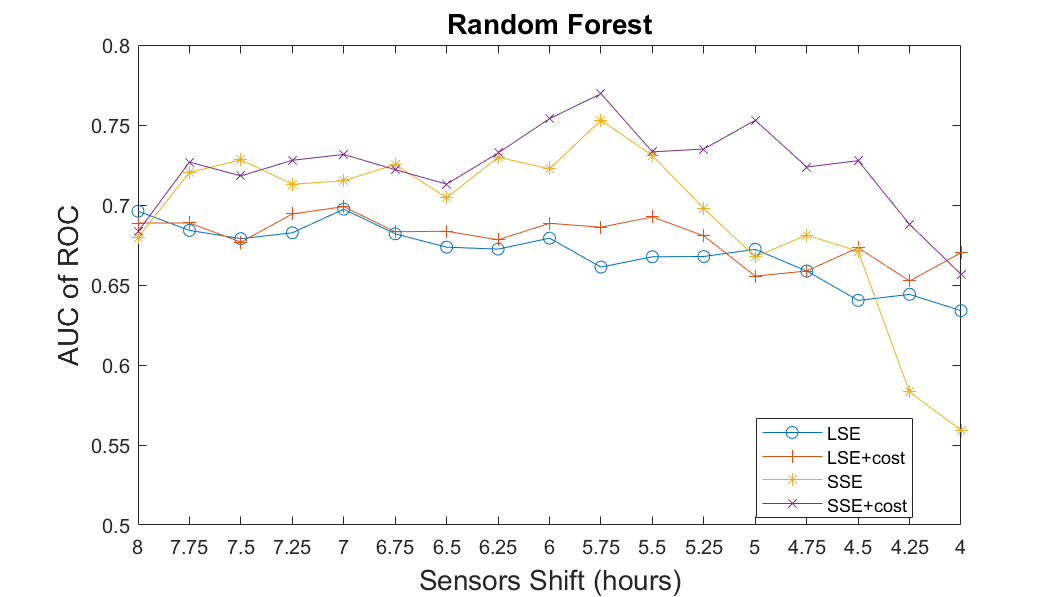}
    \caption{LSE and SSE results with RF as the base classifier.}
    \label{fig:AUC-baseDataRF}
\end{figure}

A threshold of $4$ hours is selected in both LSE and SSE because too large or too small sensors sequences may not be very useful for extracting features. Our results confirm this choice, which is described next. For both LSE and SSE, we performed classification with and without misclassification costs (as discussed in Section \ref{sec:baseline}).

\begin{figure}[!ht]
    \centering
    \includegraphics[scale=0.4]{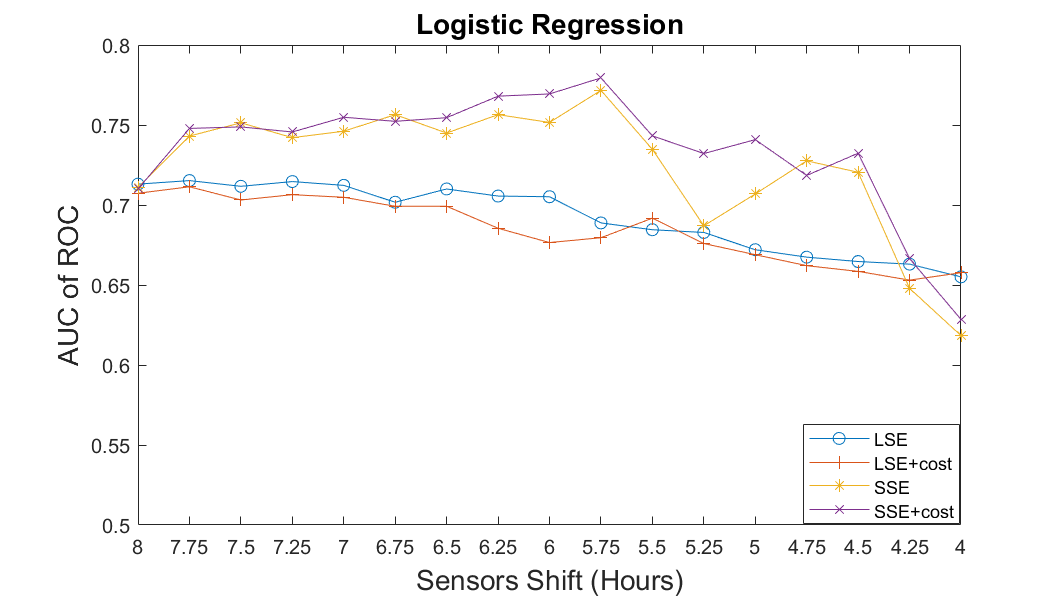}
    \caption{LSE and SSE results with LR as the base classifier.}
    \label{fig:AUC-baseDataLR}
\end{figure}

\begin{figure}[!ht]
    \centering
    \includegraphics[scale=0.4]{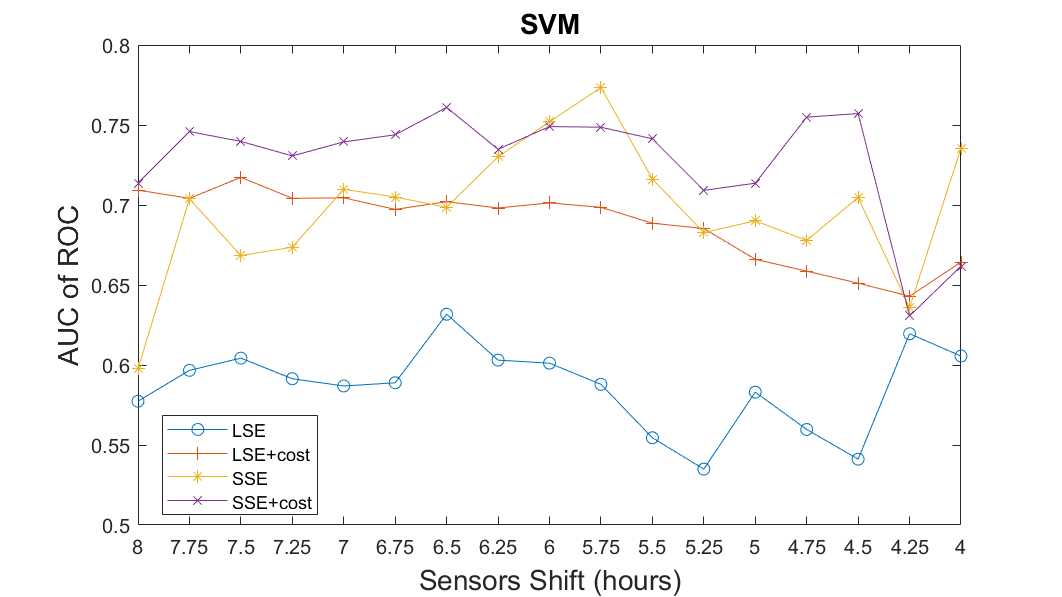}
    \caption{LSE and SSE results with SVM as the base classifier.}
    \label{fig:AUC-baseDataSVM}
\end{figure}

Figures \ref{fig:AUC-baseDataRF},  \ref{fig:AUC-baseDataLR} and \ref{fig:AUC-baseDataSVM} show the result of LSE and SSE with and without cost using RF, LR and SVM classifiers. It can be observed that SSE performed better than LSE without cost. With cost SEE performed better than LSE in all the classifiers except for one sensor shift point in SVM. This confirms our hypothesis that considering shorter sequences for feature extraction are more useful than larger sequences. The best AUC achieved with SSE without cost was $0.753$ for RF, $0.772$ for LR and $.774$ for SVM, and SSE with cost was $0.770$ for RF, $0.779$ for LR and $0.761$ for SVM. These results are a significant improvement over the the baselines AUCs reported in Section \ref{sec:baseline}.  

\begin{figure}[ht!]
    \centering
    \includegraphics[scale=0.45]{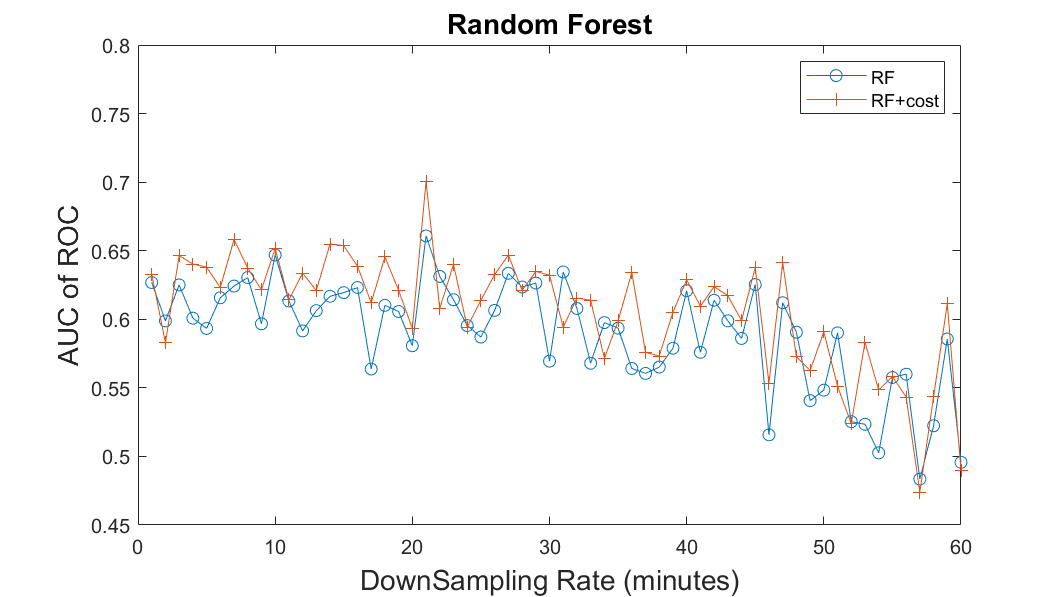}
    \caption{Performance of RF classifier at different downsampling rates (with and without cost).}
    \label{fig:AUCdsDataRF}
\end{figure}

\subsection{Sub-Sampling}

\begin{figure}[ht!]
    \centering
    \includegraphics[scale=0.45]{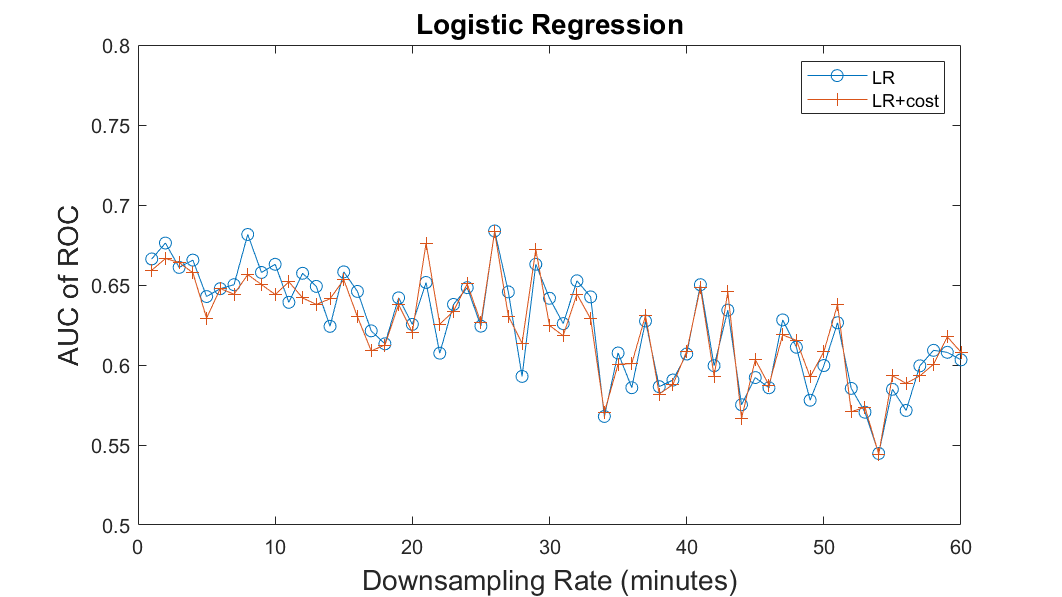}\\
    \caption{Performance of LR classifier at different downsampling rates (with and without cost).}
    \label{fig:AUCdsDataLR}
\end{figure}

\begin{figure}[ht!]
    \includegraphics[scale=0.45]{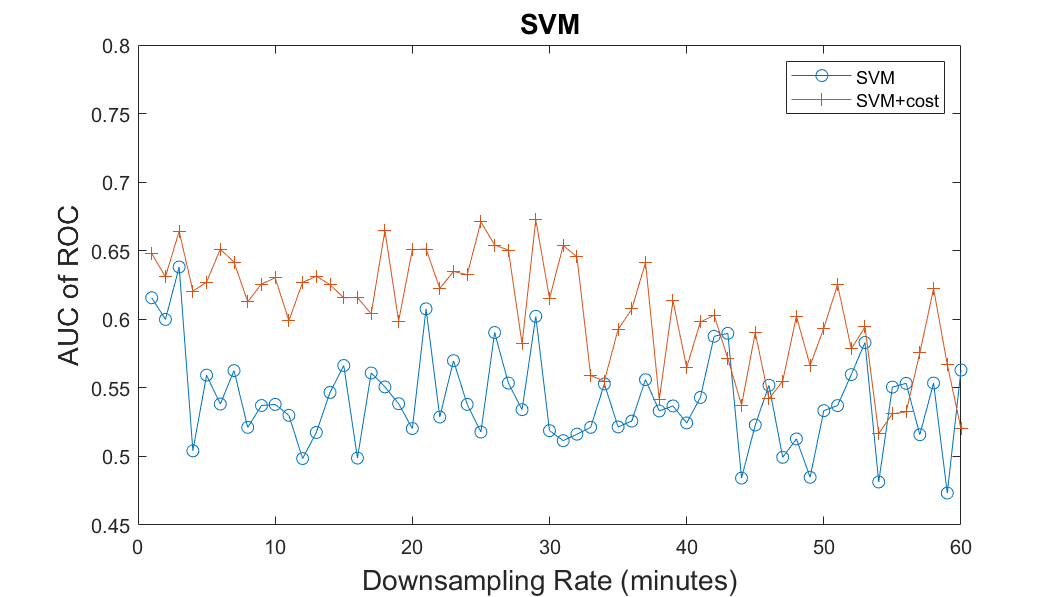}
    \caption{Performance of SVM classifier at different downsampling rates (with and without cost).}
    \label{fig:AUCdsDataSVM}
\end{figure}

One of the challenges in this dataset was the very large size of time-series / sequential multi-modal sensor data. It is very hard to extract features that are meaningful and representative of the presence or absence of agitation events in it. One possibility to reduce raw sensor data before performing feature extraction is to downsample the sensor data. However, this could also lead to loss of information due to downsampling. 
Some features described in Section \ref{sec:dataprocessing} require minimum number of points; we set this number to $5$. If downsampled data has less than $5$ points, they are discarded to avoid numerical computation problems. Therefore, as the downsampling rate increases, number of sensors shift can reduce from the original number of $693$ shifts.
To verify the performance on reduced data, we downsampled the data from $1$ minute (or $60$ seconds) to an hour (or $3600$ seconds) with an increment of $1$ minute. We did not downsample the sensor data beyond one hour because that would lead to very few points and they may not capture any useful information in the sensors data. Figures \ref{fig:AUCdsDataRF}, \ref{fig:AUCdsDataLR} and \ref{fig:AUCdsDataSVM} show the results for RF, LR and SVM classifier with and without cost when the sensor data was downsampled from $1$ minute to $60$ minutes. We observe that the cost version slightly performs better than without cost. The performance improved in each the case around $20$ minutes and then gradually decreased. It was expected that the performance would decrease at higher sampling rate due to loss of information from the raw sensor data. The interesting observation is that the best AUC on the downsampled data for each classifier ($0.700$ for RF+cost, $0.683$ for LR+cost, and $0.673$ for SVM is almost equivalent to the baselines results with full sensor data for RF (see Table \ref{tab:baseline}). This result suggests that downsampling on very long multi-modal sensor sequences where the actual timing/duration of agitation event is not known, could lead to similar results as compared to the full sensor data. Agitation events can be gradual and can extend for several minutes to hours. Therefore, despite downsampling the sensor data, useful features can still be extracted.
We could not perform a similar experiment as discussed in Section \ref{sec:improve} of choosing longer and shorter length sequences, because that value can change for every sampling rate and would lead to an increased number of hyperparameters.

\section{Conclusions and Future Work}
In this paper, we presented the results of a real-world study that collected wearable multimodal sensor data from PwD in a care environment. The task was to detect the presence or absence of agitation during a nursing shift. This is a challenging problem due to the large number of points to analyze and standard techniques (e.g. DTW) fail due to quadratic time complexity.
The time-series sequences were of unequal lengths and the actual time, duration and number of occurrences of agitation events are not available, and only one label is given at the end of very long sequence of data. We extracted features from these large multi-modal time-series data and presented a baseline results. Based on the observation that majority of sensors shifts were not equal to nursing shifts in terms of data collected, we chose shorter sequences to build models and improved the baseline results. An interesting experimental observation was that downsampling the sensor data by up to $21$ minutes gave equivalent results in comparison to full sensor data. This work also highlighted the importance of  collecting high-quality ground truth labels for clinical studies and the limitations of using clinical documentation for this purpose.
In future, we will extract localized features within a time window; however, it could also lead to variable feature length due to unequal length of sensor sequences. Encouraged by our results on downsampled data, we are currently exploring the use of  Temporal Convolution Network \cite{lea2017temporal} that can model long-range dependencies, and much faster than LSTM-based methods.

\section*{Acknowledgements}
This project was funded through the Alzheimer's Society Research Grant.



\thispagestyle{plain}
\printbibliography

\end{document}